\documentclass[10pt,twocolumn,letterpaper]{article}

\usepackage{wacv}              

\usepackage{graphicx}
\usepackage{amsmath}
\usepackage{amssymb}
\usepackage{booktabs}
\usepackage{algpseudocode}
\usepackage{algorithm}

\usepackage[pagebackref,breaklinks,colorlinks]{hyperref}

\usepackage[capitalize]{cleveref}
\crefname{section}{Sec.}{Secs.}
\Crefname{section}{Section}{Sections}
\Crefname{table}{Table}{Tables}
\crefname{table}{Tab.}{Tabs.}

\makeatletter
\DeclareRobustCommand\onedot{\futurelet\@let@token\@onedot}
\def\@onedot{\ifx\@let@token.\else.\null\fi\xspace}
\def\eg{\emph{e.g}\onedot} 
\def\ie{\emph{i.e}\onedot}

\def\etal{\emph{et al}\onedot}
\makeatother

\newcommand{\OurMethod}{SuperWeight Networks}
\newcommand{\OurMethodSpace}{\OurMethod{ }}
\newcommand{\OurMethodShort}{SWN}

\def\wacvPaperID{2192} 
\def\confName{WACV}
\def\confYear{2024}

\begin{document}

\title{Learning to Compose SuperWeights for Neural Parameter Allocation Search}


\author{Piotr Teterwak\thanks{Equal Contribution, work done while Soren was at Boston University}\\
Boston University\\
{\tt\small piotrt@bu.edu}
\and
Soren Nelson\footnotemark[1] \\
 Physical Sciences Inc\\
{\tt\small snelson@psicorp.com}
\and
Nikoli Dryden\\
ETH Z\"urich\\
{\tt\small nikoli.dryden@inf.ethz.ch}
\and
Dina Bashkirova\\
Boston University\\
{\tt\small dbash@bu.edu}
\and
Kate Saenko\\
Boston University\\
{\tt\small saenko@bu.edu}
\and
Bryan A. Plummer\\
Boston University\\
{\tt\small bplum@bu.edu}
}
\maketitle

\begin{abstract}
Neural parameter allocation search (NPAS) automates parameter sharing by obtaining weights for a network given an arbitrary, fixed parameter budget.  Prior work has two major drawbacks we aim to address.  First, there is a disconnect in the sharing pattern between the search and training steps, where weights are warped for layers of different sizes during the search to measure similarity, but not during training, resulting in reduced performance.  To address this, we generate layer weights by learning to compose sets of SuperWeights, which represent a group of trainable parameters.  These SuperWeights are created to be large enough so they can be used to represent any layer in the network, but small enough that they are computationally efficient.   The second drawback we address is the method of measuring similarity between shared parameters.  Whereas prior work compared the weights themselves, we argue this does not take into account the amount of conflict between the shared weights.  Instead, we use gradient information to identify layers with shared weights that wish to diverge from each other.  We demonstrate that our \OurMethodSpace consistently boost performance over the state-of-the-art on the ImageNet and CIFAR datasets in the NPAS setting.  We further show that our approach can generate parameters for many network architectures using the same set of weights.  This enables us to support tasks like efficient ensembling and anytime prediction, outperforming fully-parameterized ensembles with 17\% fewer parameters\footnote{Code available: \url{https://github.com/piotr-teterwak/SuperWeights}}.
\end{abstract}

\section{Introduction}

\begin{figure}[t]
    \centering
    \includegraphics[trim={0 0 0cm 0.2cm},clip,width=1.0\columnwidth]{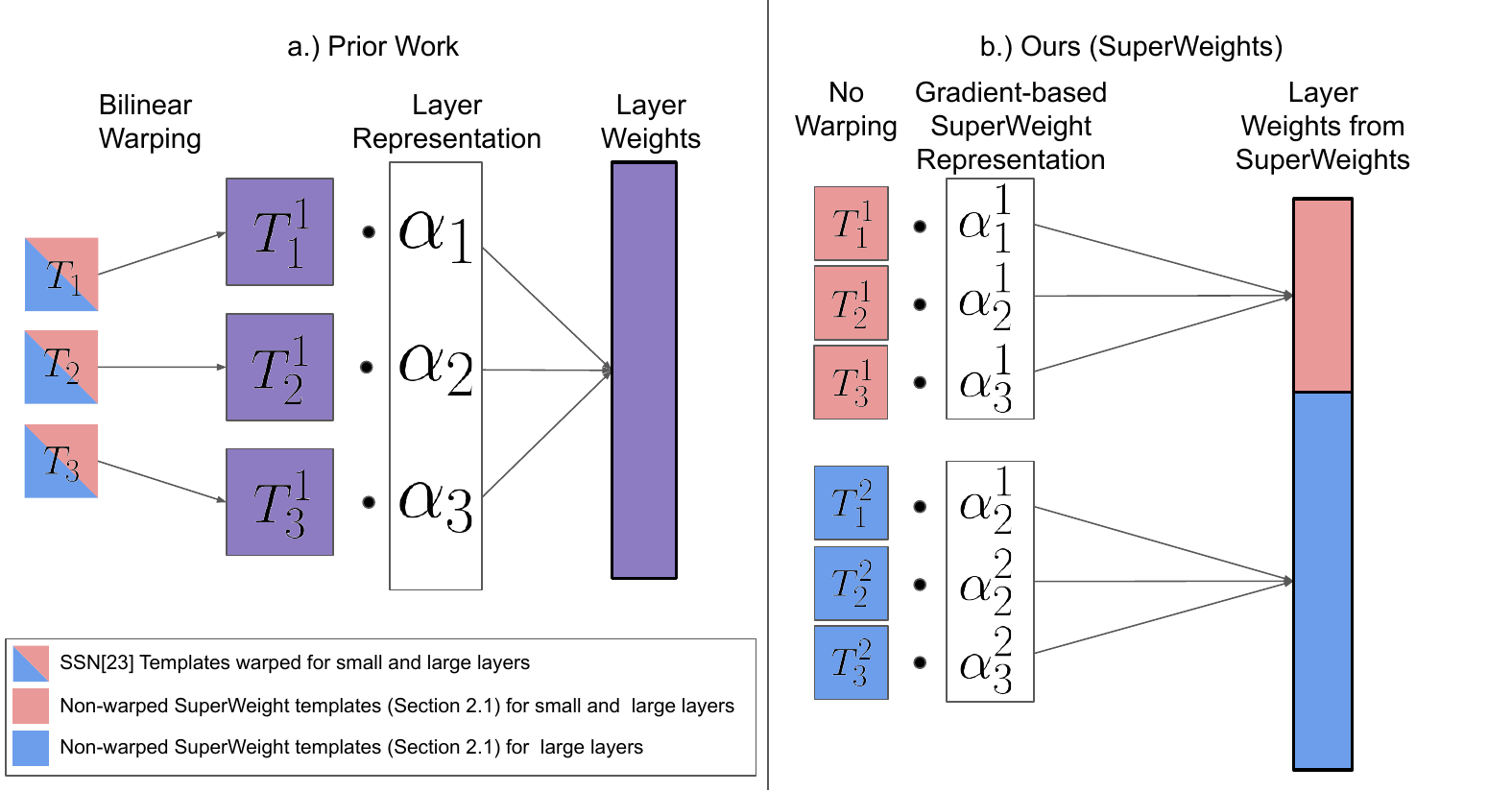}
    \caption{\textbf{Comparison to prior work. (a)} Prior work in Neural Parameter Allocation Search~\cite{plummer2020shapeshifter} would search for a good parameter sharing strategy by comparing the linear weights learned by each layer over a set of shared templates that warped to the size required by each layer using bilinear interpolation only when searching for a good parameter sharing strategy.  In contrast, \textbf{(b)} illustrates our \OurMethod, which creates SuperWeights that are concatenated together to form a layer's weights.  Each SuperWeight is only shared by layers of a minimize size, helping us to avoid warping the layers in the search step and ensuring there is no disconnect between search and training stages.
    }
    \label{fig:motivation}
\end{figure}

Parameter sharing is used to increase the computational efficiency and/or accuracy of neural networks (\eg,~\cite{bagherinezhad2017lcnn,ha2016hypernetworks,jaegle2021perceiver,Lan2020ALBERT,plummer2020shapeshifter,savarese2019learning,wallingford2022task}).  Instead of using hand-crafted heuristics to obtain a good parameter sharing strategy, Plummer~\etal~\cite{plummer2020shapeshifter} introduced a task they called Neural Parameter Allocation Search (NPAS), which automatically determines where and how to share parameters in any neural network.  A key challenge in this task is learning when sharing can occur between layers that may perform different operations (\eg, convolutional and fully connected) and have different sizes. Plummer~\etal addressed this issue by using bilinear interpolation to warp the shared parameters to the size of each layer in the network (illustrated in Figure~\ref{fig:motivation}(a)).  However, this was computationally costly as each layer would be need to be re-warped on every forward pass.  In addition, this approach also made training the network more challenging as the effective receptive field for a set of parameters would vary as its size was changed.

\begin{figure*}[t]
    \centering
    \includegraphics[width=\textwidth]{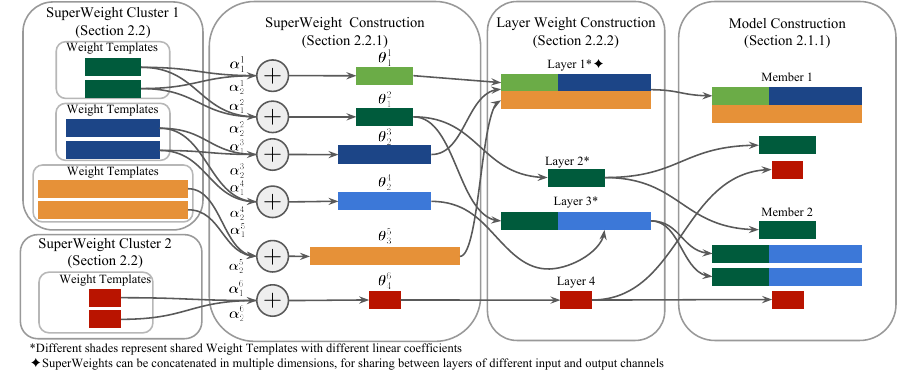}
    \caption{ \textbf{\OurMethodSpace overview.} \textbf{Left:} Weight Templates are clustered into SuperWeight Clusters; layers can only share SuperWeights from Weight Templates in the same cluster (Section \ref{subsec:learn_super}) \textbf{ Center Left:} SuperWeights are constructed from linear combinations of Weight Templates (see Eq.~(\ref{eq:mixture})) using the SuperWeight Clusters learned in Section~\ref{sec:search}. \textbf{Center Right:} The SuperWeights assigned to a layer are then concatenated to generate layer weights.  We automatically identify when we should use the same linear coefficients or different coefficients for generating a layer's SuperWeight using the procedure in Section~\ref{sec:refine}. \textbf{Right:} These layer weights are then assembled one or more ensemble members, which can have diverse architectures.}
    \label{fig:arch}
\end{figure*}

To address these issues, we present \OurMethod, a method for NPAS that effectively automates parameter sharing between diverse architectures.  A key contribution of our work is introducing the concept of a SuperWeight, which are groups of parameters that can be thought of as a feature detector that captures a unique pattern (\eg, an edge detector).  Instead of warping our parameters to fit layers of varying size as in prior work~\cite{plummer2020shapeshifter}, we create SuperWeights that are small enough to be shared between layers.  Then, as shown in Figure~\ref{fig:motivation}(b), we compose them together to create a layer's weights.  Thus, our search task transforms into locating instances where two layers find the same SuperWeight useful.

To find effective sharing strategies between our SuperWeights, we introduce a new mechanism for measuring similarity between parameters.  Specifically, Plummer~\etal~\cite{plummer2020shapeshifter} measured similarity using the value of some of the shared parameters themselves.  However, two layers could potentially have a conflict, where they both wish to alter the parameters in opposing directions, but effectively cancel their gradients out.  Such conflicts have been observed in parameter sharing settings in prior work (\eg,~\cite{yu2020gradient}), and we found the likelihood that such conflicts increases as the number of layers increase.  In other words, these networks may contain similar parameters, but the gradients from these layers would be very dissimilar.  Thus, we propose a gradient analysis approach for determining where SuperWeights can be reused.

To further improve our model's parameter efficiency, we also take advantage of template mixing methods from prior work (\eg,~\cite{bagherinezhad2017lcnn,plummer2020shapeshifter,savarese2019learning}).  We construct SuperWeights using a weighted linear combination of templates made up of trainable parameters which we call Weight Templates.  This creates a hierarchical representation for neural network weight generation, where we begin by combining Weight Templates to create SuperWeights, then concatenating together SuperWeights to create the layer weights.  This template mixing can boost performance when reusing the same parameters many times, as each combination of templates can use a unique set of coefficients. We can share parameters within a layer of a single network (Section \ref{sec:single_model}) or between different members of an ensemble (Section \ref{sec:experiments_ensemble} and \ref{subsec:anytime_results}). This allows our method to be used for efficient ensembling and anytime prediction, in addition to the NPAS task.

Our SuperWeight formulation has some similarities to Slimmable Networks~\cite{yu2019universally,yu2018slimmable}, which trained multiple networks of varying widths to support a range of inference times using a hand-crafted sharing strategy.  In contrast, our task is to automatically search for a good sharing strategy, including within a single network.  In addition, one goal of NPAS methods like ours is to support training networks with fewer computational resources, whereas Slimmable Networks use additional resources to train each subnetwork. Several other tasks also used parameter sharing to make more efficient networks, including those applied to neural architecture search (\eg,~\cite{hu2022generalizing,yu2019autoslim,yu2020bignas,zaidi2020neural,zhou2022close}), mixture-of-experts models~\cite{shazeer2017outrageously,riquelme2021scaling,wang2020deep,zhou2022Experts}, or modular and self-assembling networks~\cite{Alet2019modular,alet2018modular,devilICRA2017}.  However, these tasks are orthogonal to each other as they focus on optimizing the network architecture, whereas we aim to optimize the sharing strategy between layers.  Thus, they often can be combined.

In summary, our contributions are as follows:   

\begin{itemize}
    \item We propose \OurMethod, a new technique for automated parameter sharing across layers that composes reusable SuperWeights across layers in a network to support diverse architectures. 
    \item We introduce a new approach to computing parameter similarity that uses gradient information rather than the values of the parameters themselves, enabling us to find more effective sharing strategies that consistently improve performance over the state-of-the-art.
    \item We demonstrate that our approach is also effective at searching for sharing strategies over ensembles of diverse models.  This enables us to surpass the performance fully-parameterized ensembles on CIFAR while still using 17\% fewer parameters as well as obtain state-of-the-art performance on anytime inference.
\end{itemize}


\section{Learning to Share with \OurMethod}
\label{sec:method}

Given a parameter budget $B$ and layers $\ell_{1,...,S}$, the goal of Neural Parameter Allocation Search (NPAS) is to generate the weights of the layers of each member that maximize task performance. Plummer~\etal~\cite{plummer2020shapeshifter} only explored settings where all the layers $\ell_i$ belonged to the same network, but in this paper these layers can be separated into $M$ target architectures, each of which can be generated and make independent predictions. 
To address this task, we propose \OurMethod, an automated approach for sharing parameters with a hierarchical weight construction construction process.  At the lowest level, we perform a linear combination of Weight Templates that contain trainable parameters to generate our SuperWeights (discussed in Section~\ref{sec:generattion}).  Then we concatenate together SuperWeights to create weights for a layer (described in Section~\ref{subsec:super_size}).  An overview of this process is illustrated in Figure~\ref{fig:arch}.

To find a good parameter sharing strategy, we use a two step search-and-refine process. This begins by identifying what layers may share trainable parameters effectively, which we group together into SuperWeight Clusters (Section~\ref{sec:search}).  Then we determine what layers can share SuperWeights effectively (Section~\ref{sec:refine}).  In both these stages we will leverage gradient information to measure how similar the parameters are (and, therefore, how much they would benefit from parameter sharing).

\subsection{Generating a Neural Network using SuperWeights}
\label{sec:banks2superweights}

In this section we shall describe how to generate weights for a layer in a neural network, which will provide context for our search-and-refine procedure for creating our SuperWeight Clusters in Section~\ref{subsec:learn_super}. Traditional methods of (hard) parameter sharing directly reuse weights  (\eg,~\cite{collobert2008unified,havasi2020training,gal2016dropout,zhang2014facial}), but this limits layer weight diversity as shared layers then encode the same learned function.   Instead, we take inspiration from recent work in cross-layer parameter sharing (\eg,~\cite{bagherinezhad2017lcnn,plummer2020shapeshifter,savarese2019learning}) that performs template mixing. In these works, parameters and layer weights are decoupled; each layer's weights are a linear combination of parameter matrices called templates.

\subsubsection{Network Generation Process}
\label{sec:generattion}

Consider our generation process outlined in Figure~\ref{fig:arch}, which we shall step through from right-to-left.  In the rightmost column describing model construction, we see seven layers across two different networks.  The first layer of ``Member 1'' is generated by concatenating together three SuperWeights (labeled as ``Layer Weight Construction'' in the third column).  These SuperWeights are generated by a linear combination of Weight Templates (``SuperWeight Construction''), whereas the Weight Templates may share parameters between any SuperWeights within the same SuperWeight Cluster.  

Now going from left-to-right through the generation process: our first step is to create a set of Weight Templates $T^g_{1,...,N}$ with the same dimensions as the $kth$ SuperWeight $\theta^{(k)}_g$.  Following~\cite{plummer2020shapeshifter}, these Weight Templates are created by splitting the trainable parameters into templates and allocating them to SuperWeights in a round robin fashion.  Then we generate our SuperWeight $\theta^{(k)}_g$ via a linear combination using coefficients $\alpha^{(k)}$, \ie,
\begin{equation}
\label{eq:mixture}
 \theta^{(k)}_g = \sum_{i=1}^{N} \alpha^{(k)}_i T^g_i
\end{equation}
Multiple SuperWeights can share $g$ while having different coefficients $\alpha^{(k)}$ (represented as different color shades for a SuperWeight, \eg, green, in the second column of Figure~\ref{fig:arch}). To create a layer's weights (third column of Figure~\ref{fig:arch}), one or more SuperWeights are generated using Eq.~(\ref{eq:mixture}) and then concatenated together (additional details in Section~\ref{subsec:super_size}). These layers are stacked together to create a neural network (last column of Figure~\ref{fig:arch}). \emph{Note that templates and coefficients are learned jointly via gradient descent}.

Plummer~\etal~\cite{plummer2020shapeshifter} also shared parameters via template mixing between layers of different sizes, but they would directly reshape the available parameters into templates of the target layer size.  This means that it would be very difficult to accurately represent subnetworks within a larger network.  For example, let us assume we are generating the weights for two networks that are identical except that network $X$ is twice as wide as network $Y$.  The approach from Plummer~\etal would either have to generate layers for $X$ and $Y$ independently, or we would generate the layers of $X$, and then slice out the weights that would fit network $Y$.  We found that the first option was difficult to optimize that resulted in lower performance in our experiments.  The second option would reduce diversity, as network $X$ would contain exactly the same weights as network $Y$.  Instead, our \OurMethodSpace can directly optimize any subnetworks, like in the second option in our example above, but while minimizing any diversity loss (details Section~\ref{subsec:learn_super}).

\subsubsection{Determining SuperWeight Sizes}
\label{subsec:super_size}

The target sizes for our SuperWeights are based on layer shapes. Consider the case where we are generating weights for a set of layers $\ell_{1,...,S}$, where each layer is a different size.  We create an initial set of SuperWeights by sorting the layers in order from smallest to largest.  The first SuperWeight we would create would be the same size as the smallest layer $\ell_i$, and every layer within that SuperWeight Cluster that was the same size as $\ell_i$ would initially share the same SuperWeight.  Then, when we generate the weights for the next layer $\ell_{i+1}$, we assume we have the SuperWeight from layer $\ell_i$.  Thus, the SuperWeights for layer $\ell_{i+1}$ need only generate the additional weights required beyond those provided by the SuperWeights from $\ell_i$.  

For example, if $\ell_i$ needed 100K parameters and $\ell_{i+1}$ needed 400K, then the new SuperWeights would generate $400-100=300$K weights. If $\ell_{i+1}$ is larger than $\ell_{i}$ in only a single dimension, then only a single new SuperWeight is generated. If it's larger in both input and output channels, $\ell_{i+1}$ generates two new SuperWeights. The first SuperWeight extends the size in a input channels dimension, and the second is concatenated in the output channel dimension. See Layer 1 in the third column of Figure \ref{fig:arch} for a graphical illustration. Any subsequent layers would follow the same process, but would have all the weights from the previous layers (\ie, $\ell_{i+2}$ would have 400K weights from $\ell_{i+1}$, 100K of which originally came from $\ell_{i}$).

\subsection{Finding Where to Share via Search-and-Refine with Gradient Similarity}
\label{subsec:learn_super}

There are two components of Figure~\ref{fig:arch} that we did not discuss in the network generation details found in Section~\ref{sec:banks2superweights}.  First, our SuperWeight Cluster creation process, which separates SuperWeights into groups that share trainable parameters (shown in the first column of the figure).  In this setting we find what layers can effectively share trainable parameters (described in Section~\ref{sec:search}).  This can be thought of as trying to learn a coarse policy where we identify layers that find any sharing detrimental.  

After Plummer~\etal~\cite{plummer2020shapeshifter} found a good sharing strategy, they would train their model using soft-sharing, where every layer has their own coefficients $\alpha^{(k)}$ when performing a combination of templates (\eg, similar to using Eq.~\ref{eq:mixture}).  However, our approach differs in that we consider the case that some SuperWeights may find hard-sharing across layers beneficial (\eg, the dark green SuperWeight being reused in Layers 2 and 3 in Figure~\ref{fig:arch}).  Thus, after we begin training our model, we first perform hard-sharing across all layers in the the same SuperWeight Cluster (\ie, the same set of coefficients $\alpha^{(k)}$ are used for every SuperWeight).  After a few epochs, we refine our sharing policy so that layers that find hard-sharing challenging can begin using soft-sharing instead, \ie, they would begin using their own coefficients, but still share Weight Templates, as illustrated with the two SuperWeights of varying blue shades in Figure~\ref{fig:arch} (described in Section~\ref{sec:refine}).

\subsubsection{Searching for a Coarse Parameter Sharing Policy}
\label{sec:search}

Our first step in finding a good sharing strategy is to determine what layers can effectively share parameters to create Superweight Clusters.  This provides our coarse sharing policy that we shall refine in Section~\ref{sec:refine}.  There are two key differences with our search step from prior work~\cite{plummer2020shapeshifter}.  First, our SuperWeight construction allows us to avoid warping the shared templates as discussed in the Introduction, meaning that how we will perform the search step will be using the same mechanisms as when we train our model.  Second, rather than using the coefficients learned by each layer $\alpha^{(k)}$ to measure similarity between parameters, we measure similarly by comparing the gradients a SuperWeight is receiving from the layers that share it.  The intuition behind this approach is that layers with conflicting gradients are playing a gradient tug-of-war, hurting optimization. In other words, if the sum of gradients from two layers is close to zero, no learning occurs.

More formally, given a set of layers $\ell_{1,...,S}$, our goal is to determine which layers can share parameters.  At this stage, we treat all layers are belonging to the same SuperWeight Cluster and are using hard-sharing of SuperWeights across layers.  This means that every set of Weight Templates has exactly one set of coefficients.  For example, if a cluster has three layers that all need a SuperWeight of 100K dimensions, then each layer would use a shared Weight Template that produces the same 100K dimension SuperWeight (additional details found in Section~\ref{subsec:super_size}).

Thus, we can infer that if the gradients of the loss w.r.t.\ the layers sharing SuperWeights are misaligned, then the model will have optimization difficulties. Thus, we compute gradient similarity over SuperWeights to determine what layers may be grouped.  However, some layers may be composed of multiple SuperWeights, only some of which may share with other layers in the initial cluster.  For example, Layer 2 in the third column of Figure~\ref{fig:arch} has one SuperWeight, but Layer 3 has two SuperWeights (only one of which is shared with Layer 2). $\upsilon_{i,j}$ refers to the set of SuperWeights shared by layers $i$ and $j$. $W_i$ refers to the instantiated weights of layer $i$. Then the gradient similarity between $W_i$ and $W_j$ would be computed as: 
\begin{equation}
    \psi_{SW} = \cos\left(\frac{\partial \mathcal{L}}{\partial W_i}\frac{\partial W_i}{\partial \upsilon_{i,j}}, \frac{\partial \mathcal{L}}{\partial W_j}\frac{\partial W_j}{\partial \upsilon_{i,j}} \right) > \tau,
    \label{eq:sw_sim}
\end{equation}
where $\tau$ is a hyperparameter representing the minimum threshold for which two layers that share $\upsilon_{i,j}$ should remain in the same cluster, and $\mathcal{L}$ is the loss.

After training for a few epochs, we create our SuperWeight Clusters by placing layers sharing the same SuperWeight into a priority queue by the cosine similarity with respect to the gradients of the shared coefficients aggregated over an epoch.  We pop the first two layers, $\ell_i,\ell_j$ off the queue and check whether their cosine similarity exceeds the threshold $\tau$, \ie, it satisfies Eq.~\ref{eq:sw_sim}.  If they do not satisfy Eq.~\ref{eq:sw_sim}, we split any ungrouped layers into their own individual groups (\eg, 4 ungrouped layers would result in 4 additional single layer groups).  Otherwise, if they do Eq.~\ref{eq:sw_sim} satisify and both layers belong to an existing group, we merge their groups. If only one layer is in an existing group, then we add the new layer into that group's set. The final set of layer groups is referred to as our SuperWeight Cluster. 
We set $\tau$ via grid search on a validation set (we found $\tau=0.1$ worked well in our experiments). Please refer to the supplementary for pseudocode for our priority-queue assignment procedure. Following~\cite{plummer2020shapeshifter}, we reinitialize our model and re-train from scratch with our new SuperWeight Clusters, where we refine our sharing policy as described in the next section.

\subsubsection{Learning when SuperWeights should share Weight Template coefficients}
\label{sec:refine}

After we obtain a set of SuperWeights Clusters from Section~\ref{sec:search}, we start training the network by only using hard-parameter sharing between SuperWeights. In other words, each layer that shares a SuperWeight will begin by using the same set of coefficients $\alpha^{(k)}$ used to combine weight templates for the first $E$ epochs of training (we found $E=10$ worked well).  After that, we analyze which layers may benefit from refining the sharing policy by decoupling linear coefficients combining Weight Templates.  In this way, unlike Plummer~\etal~\cite{plummer2020shapeshifter} which only used soft-sharing, we allow a combination of hard and soft-sharing.  

To search for a refined sharing policy, we use the same priority-queue assignment procedure as described in Section~\ref{sec:search}.  In other words, we place the layers sharing the same SuperWeight into a priority queue.  Then, we iterate over pairs of layers $\ell_i,\ell_j$ and check whether their cosine similarity exceeds some threshold $\beta$, \ie,
\begin{equation}
    \psi_{coef.} = \cos \left(\frac{\partial \mathcal{L}}{\alpha^{k,j}}, \frac{\partial \mathcal{L}}{\alpha^{k,i}} \right) > \beta,
    \label{eq:co_sim}
\end{equation}
where $\alpha^{k,i}$ is the shared coefficient corresponding to layer $\ell_i$.  Layers that satisfy Eq.~\ref{eq:co_sim} will be grouped together.  Any layers that do not satisfy Eq.~\ref{eq:co_sim} with any other layer will remain in their own group.  After we have identified the groups of similar layers, each obtains their own copy of the coefficients for that SuperWeight $\alpha^{(k)}$ and we resume training.  This creates a new SuperWeight for each group of layers whose gradients point in a similar direction, allowing for layer specialization.

\section{Single Network Search Results}
\label{sec:single_model}

\begin{table*}[t]
\centering
\setlength{\tabcolsep}{4pt}
    \begin{tabular}{rlcccccc} 
 \toprule
  & & \multicolumn{2}{c}{CIFAR-10~\cite{cifar100}} & \multicolumn{2}{c}{CIFAR-100~\cite{cifar100}}  & \multicolumn{2}{c}{ImageNet~\cite{deng2009imagenet}}\\ 
  & Param Budget \% & 1\% & 10\% & 1\% & 10\% & 5\% & 10\% \\
 \midrule
  \textbf{(a)} & Baseline \cite{zagoruyko2016wide} &  93.54{\footnotesize $\pm 0.19$} &  95.73{\footnotesize $\pm 0.17$} & 71.49{\footnotesize $\pm 0.29$} & 77.43{\footnotesize $\pm 0.24$} & 66.89{\footnotesize $\pm 0.22$} & 68.12{\footnotesize $\pm 0.18$}\\
 & Single Cluster &  93.39{\footnotesize $\pm 0.17$} & 95.56{\footnotesize $\pm 0.10$} & 71.30{\footnotesize $\pm 0.26$} & 77.19{\footnotesize $\pm 0.29$} &  66.27{\footnotesize $\pm 0.36$} & 67.55{\footnotesize $\pm 0.28$}\\
 & Random Cluster & 93.48{\footnotesize $\pm 0.31$}  & 95.62{\footnotesize $\pm 0.11$} & 70.97{\footnotesize $\pm 0.50$} & 76.82{\footnotesize $\pm 0.40$} &  66.81{\footnotesize $\pm 0.41$} & 67.95{\footnotesize $\pm 0.43$}\\
 & SSN~\cite{plummer2020shapeshifter}  & 94.74{\footnotesize $\pm 0.16$} & 95.83{\footnotesize $\pm 0.10$} & 74.66{\footnotesize $\pm 0.29$} & 78.17{\footnotesize $\pm 0.27$} &  67.69{\footnotesize $\pm 0.19$} & 70.39{\footnotesize $\pm 0.22$}\\
 \midrule
 \textbf{(b)} &\OurMethodShort~w/o Grad Sim & 94.80{\footnotesize $\pm 0.11$} & 95.84{\footnotesize $\pm 0.12$} & 74.99{\footnotesize $\pm 0.24$} & 78.57{\footnotesize $\pm 0.23$} &  68.24{\footnotesize $\pm 0.19$} &  70.79{\footnotesize $\pm 0.13$}\\
 &\OurMethodShort~w/o Refine & 94.81{\footnotesize $\pm 0.13$} & 95.95{\footnotesize $\pm 0.14$} & 75.49{\footnotesize $\pm 0.20$} & 78.26{\footnotesize $\pm 0.21$} &  68.11{\footnotesize $\pm 0.23$} &  70.69{\footnotesize $\pm 0.16$}\\
& \OurMethodShort~(\textbf{Ours}) & \textbf{94.87}{\footnotesize $\pm 0.14$} & \textbf{95.99}{\footnotesize $\pm 0.11$} & \textbf{75.77}{\footnotesize $\pm 0.34$} & \textbf{78.94}{\footnotesize $\pm 0.26$} & \textbf{68.42}{\footnotesize $\pm 0.21$} & \textbf{71.14}{\footnotesize $\pm 0.18$}\\
\bottomrule
\end{tabular}
\caption{Comparing methods that searching for parameter sharing strategies using a WRN 28-10~\cite{zagoruyko2016wide} for CIFAR and WRN 50-2 for ImageNet averaged over three runs.  Baseline reduces the width/number of layers to support a given parameter budget, which is reported as percentage of the original model's parameters. \textbf{(a)} reports the performance of our baseline methods from prior work reproduced using the author's code. \textbf{(b)} contains ablations of our \OurMethodShort, where we report a consistent boost over the state-of-the-art.  See Section~\ref{sec:single_model} for discussion.}
   \label{tab:npas_results} 
\end{table*}


We first compare our \OurMethodSpace (\OurMethodShort) to prior work on the Neural Parameter Allocation Search (NPAS) task in Section~\ref{sec:single_model}, whereas in Section~\ref{sec:experiments_ensemble} we will explore a new setting where we apply our NPAS approach to tasks that can be implemented using an ensemble of models that we generate from a set of shared parameters. 
\smallskip


\noindent\textbf{Datasets and metrics.} We evaluate our method on three standard benchmarks: CIFAR-10~\cite{cifar100}, CIFAR-100~\cite{cifar100}, and ImageNet~\cite{deng2009imagenet}. We evaluate the performance of a model based on its top-1 accuracy given a parameter budget.  Additional details on our experimental setup and hyperparameters can be found in the supplementary material.
\smallskip


\subsection{Results} 
Table~\ref{tab:npas_results} compares different strategies for creating a parameter sharing strategy for image classification.  Table~\ref{tab:npas_results}(a) reports the performance of prior work reproduced using the author's code.  Comparing the last lines of Table~\ref{tab:npas_results}(a) representing the state-of-the-art NPAS approach~\cite{plummer2020shapeshifter} and Table~\ref{tab:npas_results}(b) that reports our full method, we see we obtain a consistent boost over prior work.  Notably, we receive around a 1\% boost on the ImageNet dataset.  Overall, we find that when a sharing strategy is most needed, \ie, in the lower budget settings, we observe a larger boost to performance (increasing to an almost 2\% gain on ImageNet).

When comparing the effect of the different components of our \OurMethodSpace in Table~\ref{tab:npas_results}(b), find that each contributes meaningfully to the final model performance.  The gradient similarity function provides the most benefit on the ImageNet dataset, where we see a 0.5\% gain to top-1 accuracy. In addition, also compare to using only perform the search step from Section~\ref{sec:search}, but skip the refinement step in Section~\ref{sec:refine}, (similar to the search strategy of SSNs~\cite{plummer2020shapeshifter}).  Another way of thinking of this comparison is measuring the effect of hard-sharing SuperWeights across layer (w/o Refine) vs.\ using a mix of hard and soft sharing in our full model.  Comparing the second and third lines of Table~\ref{tab:npas_results}(b), we see that the refinement step is key to good performance in some settings, \eg, we see more than a 0.5\% gain on CIFAR-100 with a 10\% budget. In the next section, we will explore applications of our work where we are tasked with finding a good parameter sharing strategy across an ensemble of models and architectures.

\begin{table*}[t]
\centering
\setlength{\tabcolsep}{4.pt}
 \begin{tabular}{rlcccccccccc} 
 \toprule
  & & & \multicolumn{3}{c}{CIFAR-100 (clean)} & \multicolumn{3}{c}{CIFAR-100-C} & \multicolumn{3}{c}{CIFAR-10}\\ 
 & Method & Params & Top-1 $\uparrow$ &  NLL $\downarrow$ &  ECE $\downarrow$ & Top-1 $\uparrow$ & NLL $\downarrow$ & ECE $\downarrow$ & Top-1 $\uparrow$ & NLL $\downarrow$ & ECE $\downarrow$\\
 \midrule
  \textbf{(a)} & {WRN-28-10}& 36.5M & 79.8 & 0.875 & 8.6 & 51.4 & 2.70 & 23.9 & 96.0 & 0.159 &  2.3\\
 & {BE~\cite{wen2020batchensemble}}& 36.5M & 81.5 & 0.740  & 5.6 & \textbf{54.1} & 2.49 & 19.1 & 96.2 & 0.143 & 2.1\\
 & {BE~\cite{wen2020batchensemble} + EnsBN} & 36.5M & 81.9 & n/a & 2.8 & \textbf{54.1} & n/a & 19.1 & 96.2 & n/a & 1.8\\
 & {MIMO~\cite{havasi2020training}  }& 36.5M & 82.0 & 0.690  & 2.2 & 53.7 & 2.28 & 12.9 & 96.4 & 0.123 & 1.0\\
 & {Thin Deep Ensembles} & 36.5M & 81.5 & 0.694 & \textbf{1.7} & 53.7 & 2.19 & 11.1 & 96.3 & \textbf{0.115} & \textbf{0.8}\\
 & {\OurMethodShort-HO (\textbf{Ours})} & 36.5M & 82.2 & 0.702 & 2.7& 52.9 & \textbf{2.17} & 10.3& 96.3 & 0.120 & \textbf{0.8}\\
 & {\OurMethodShort-HE (\textbf{Ours})} & 36.5M & \textbf{82.4} & \textbf{0.663} & 3.0 & 53.0 & \textbf{2.17} & \textbf{10.0} & \textbf{96.5} & \textbf{0.115} & \textbf{0.8}\\
 \midrule
  \textbf{(b)} & {Deep Ensembles } & 146M & 82.7 & \textbf{0.666} & \textbf{2.1} & 54.1 & 2.27 & 13.8 & \textbf{96.6} & \textbf{0.114} &  1.0\\
  & {\OurMethodShort-HO (\textbf{Ours})}& 120M &\textbf{82.9} &\textbf{0.666} &2.2&\textbf{54.7} &\textbf{2.00} & \textbf{10.3} &\textbf{96.6} &0.119&\textbf{0.8}\\
  \bottomrule
\end{tabular}
\caption{\textbf{Homogeneous ensembling comparison} on CIFAR-100 (clean)~\cite{cifar100} CIFAR-100-C (corrupt)~\cite{hendrycks2019robustness}, and CIFAR-10 using  WideResNets~\cite{zagoruyko2016wide} averaged over three runs. \textbf{(a)} shows that our approach outperforms prior work in efficient ensembling \cite{wen2020batchensemble,havasi2020training}.  \textbf{(b)} compares the performance of increasing the number of parameters (without changing the architecture) using our approach compared to Deep Ensembles, which trains 4 independent networks as members. 
}
\label{table:cifar100}
\vspace{-4mm}
\end{table*}

\section{Multi-Network Search Experiments}
\label{sec:experiments_ensemble}
In this paper we explore a new application of NPAS methods, where rather than searching for a good parameter sharing strategy over a single network, they must search over multiple architectures.  We find that this work spans two different application areas: efficient ensembling (\eg, \cite{lee2015m,wen2020batchensemble,wenzel2020hyperparameter}) and anytime inference (\eg,~\cite{ruiz2021anytime,havasi2020training,wen2020batchensemble,yu2018slimmable,yu2019autoslim}).  In efficient ensembling the goal is to reduce the computational resources required to support a ensemble of models.  These often make strong architectural assumptions, such as ensemble member homogeneity (\ie, each member is the same architecture), which limits their use. For example, homogeneous ensembles are ill-suited to tasks like anytime prediction because one only has $n$ options for computational complexity, where $n$ is the number of ensemble members. In contrast, heterogeneous ensembles can select a subset of its ensemble members to provide a range of inference times (\eg, a 4 member heterogeneous ensemble can adjust to ${4 \choose 1} + {4 \choose 2} + {4 \choose 3} + {4 \choose 4} = 15$ levels of inference latency).  We demonstrate the flexibility and generalization power of our approach by addressing both tasks. Implementation details can be found in the supplementary.
\smallskip

\noindent\textbf{Additional datasets and metrics.} , We  evaluate the robustness of our method on out-of-domain samples in addition to CIFAR-10 and CIFAR-100.  Specifically, we report performance on CIFAR-100-C \cite{hendrycks2019robustness}, which is the CIFAR-100 test set  corrupted by distortions such as Gaussian blur and JPEG compression.  We also supplement top-1 accuracy with calibration metrics~\cite{guo2017calibration}: Negative Log-Likelihood (NLL) and Expected Calibration Error (ECE).


\subsection{Efficient Ensembling Results}
\label{sec:efficient}

Table~\ref{table:cifar100}(a) compares our \OurMethodSpace (\OurMethodShort-HO) on the CIFAR-100 CIFAR-100-C, and CIFAR-10 with prior work in efficient ensembling. Note that all the methods boost performance over a single model without requiring additional model parameters.  However, our \OurMethodSpace outperforms all other methods on CIFAR-100 when using 36.5M parameters.  Table~\ref{table:cifar100}(a) also shows that heterogeneous \OurMethodSpace Ensembles (\OurMethodShort-HE), consisting of a WRN-34-8, 28-12, 28-10, and 28-8, outperforms \OurMethodShort-HO in over half of the metrics while also supporting many inference times. 

Unlike methods like BatchEnsemble (BE) \cite{wen2020batchensemble} and MIMO \cite{havasi2020training}, which cannot change the number of parameters without making architecture adjustments to the widths and/or number of layers, our \OurMethodSpace can support any parameter budget without requiring architecture changes by adjusting the number of templates or the amount of sharing between layers. Thus, if the number of parameters are not a concern, our approach can increase our parameter budget to boost performance.  This is illustrated in Table~\ref{table:cifar100}(b), where we outperform standard Deep Ensembles, which trains independent networks for ensemble members, while still retaining 17\% fewer parameters.  
\smallskip

\noindent\textbf{Computational resources comparison.} In addition to the number of parameters that we report in Table~\ref{table:cifar100}, the training time and inference time is also a key contributor to an efficient ensembling approach.  Thus, our results for \OurMethodShort-HO in Table~\ref{table:cifar100}(a) reduces inference time by using a smaller network (WRN-28-5, the same size used by Thin Deep Ensembles) so it has a similar inference time compared with work like MIMO~\cite{havasi2020training}, which uses a WRN-28-10.  For a fair comparison to MIMO and BE, we also normalized our experiments by training time (using their learning schedules). 


\begin{figure*}[t]
    \centering
     \includegraphics[width=0.98\textwidth]{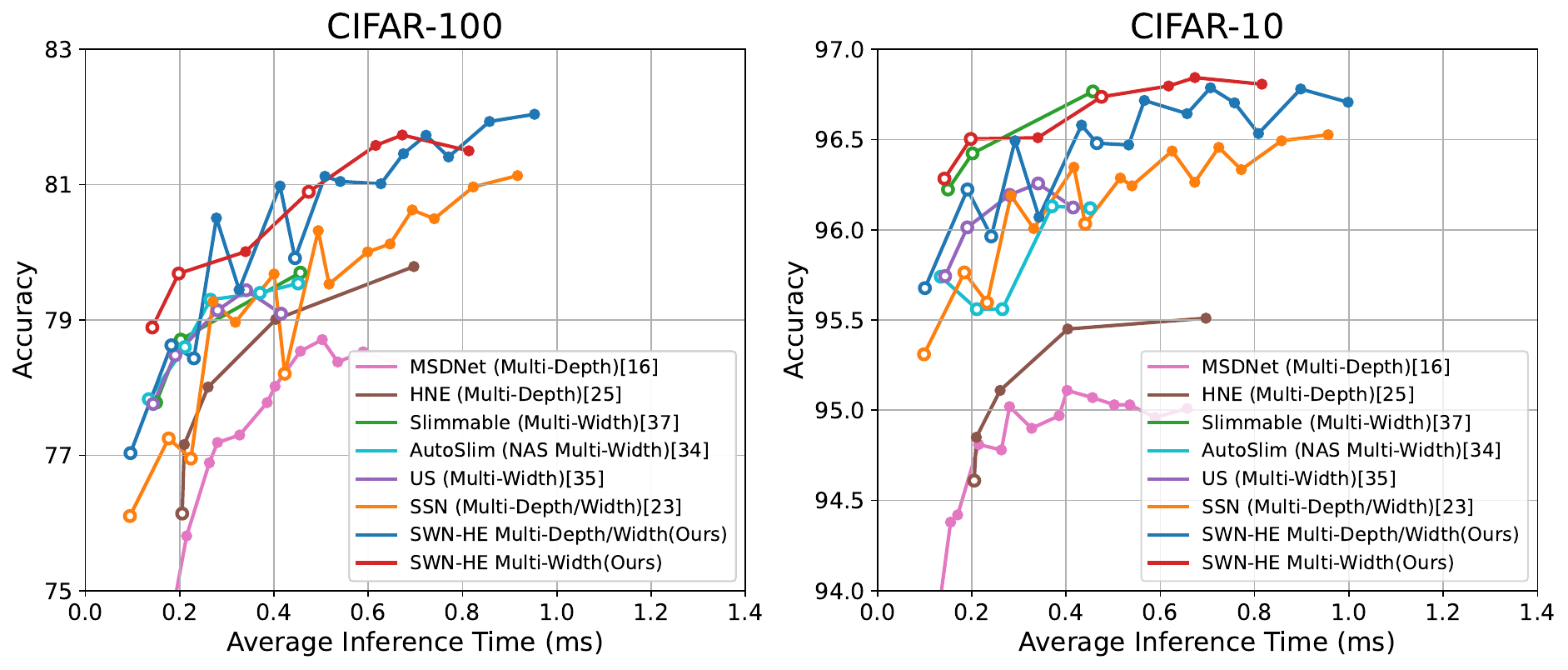}
     \caption{\textbf{Anytime Inference comparison} using CIFAR-100 and CIFAR-10~\cite{cifar100} top-1 accuracy  vs.\ inference time (ms) averaged over 3 runs on a single P100 GPU. Most methods~\cite{huang2017multi,plummer2020shapeshifter,yu2019autoslim,yu2018slimmable,yu2019universally} use WRN backbones~\cite{zagoruyko2016wide}, except HNE \cite{ruiz2021anytime} which modifies ResNet-50~\cite{he2016deep}. }
    \label{fig:anytime_cifar}
\end{figure*}
\begin{table*}[t]
\centering
 \begin{tabular}{l ccc c} 
 \toprule
 Method & WRN-28-3 & WRN-28-4 & WRN-28-7 & Full Ensemble
  \\ 
 \midrule
 Shared Coefficients              & 77.3 $\pm$ 0.09    & 78.0  $\pm$ 0.15     & 79.7       $\pm$ 0.10  & 80.4   $\pm$ 0.06     \\
Single SuperWeight Cluster & 76.3   $\pm$ 0.37   & 77.6  $\pm$ 0.30      & 78.8 $\pm$ 0.20         & 81.1   $\pm$ 0.17      \\
 {Depth-binning} & 76.8 $\pm$ 0.15 & 77.7 $\pm$ 0.17 & 79.2 $\pm$ 0.19 & 81.4 $\pm$ 0.15 \\
 {Coefficient Clustering~\cite{plummer2020shapeshifter}} & 76.1 $\pm$ 0.19 & 77.0 $\pm$ 0.15 & 78.9 $\pm$ 0.15 & 80.7 $\pm$ 0.14 \\
 {\OurMethodShort-HE (\textbf{Ours})} & \textbf{78.9 $\pm$ 0.26} & \textbf{79.7 $\pm$ 0.08} & \textbf{80.9 $\pm$ 0.03} & \textbf{81.5 $\pm$ 0.13}  \\ 
 \bottomrule
\end{tabular}
\caption{\textbf{Multi-width SuperWeight Cluster creation } using top-1 accuracy on CIFAR-100 over three runs. See Section~\ref{subsec:anytime_results} for discussion.}
\label{table:cluster_assignment}
\end{table*}

\subsection{Anytime Inference Experiments}
\label{subsec:anytime_results}

In anytime inference the goal is to make high-performing predictions within a given time budget.  As the time budget increases, a good method will use the additional time to improve performance.  A homogeneous ensemble, like those used in the experiments in Section~\ref{sec:efficient}, would only provide limited time budgets as each member has an identical computational complexity. Thus, their  flexibility is limited because one has only $M$ options for computational complexity, where $M$ is the number of ensemble members.  However, for a heterogeneous ensemble this limitation is removed with an ensemble since each member provides a different inference time resulting in ${M \choose 1} + {M \choose 2} + ... + {M \choose M}$ possible inference times to choose from. This results in a highly effective anytime inference model. We note that although we evaluate our individual ensemble members in series,  our method is trivial to parallelize to increase inference speeds by using multiple GPUs. This is unlike many early-exit anytime inference methods (\eg,~\cite{pmlr-v70-bolukbasi17a,huang2017multi,kaya2019shallow,Li2019ImprovedTF,teerapittayanon2016branchynet,yang2020resolution}), which are intrinsically serial. 

We use two settings for our Heterogeneous \OurMethodSpace (\OurMethodShort-HE) in our experiments: \OurMethodShort-HE -Multi-Width, which trains a three member WRN~\cite{zagoruyko2016wide} ensemble WRN-28-[7,4,3], and \OurMethodShort-HE-Multi-Depth/Width, a four member ensemble WRN-28-[7,4] 16-[7,4]. SSNs~\cite{plummer2020shapeshifter} and Slimmable~\cite{yu2018slimmable} use the same ensemble configurations. Other approaches use method-specific strategies (\eg, HNE~\cite{ruiz2021anytime} trains a set of tree-nested ensembles).  See supplementary for additional details, including experiments where we share parameters across different architecture families.

\subsubsection{Anytime Inference Results} 

Figure \ref{fig:anytime_cifar} reports top-1 accuracy vs.\ average inference time using a single P100 GPU on CIFAR-10 and CIFAR-100~\cite{cifar100}. When comparing to other dynamic width methods such as Slimmable~\cite{yu2018slimmable}, Universally Slimmable~\cite{yu2019universally} and AutoSlim~\cite{yu2019autoslim} models, \OurMethodSpace perform on par or better than them for inference times they support, but our approach can provide a wider range of inference times that improve performance. We reiterate that other efficient ensembling methods such as BatchEnsemble \cite{wen2020batchensemble} and MIMO \cite{havasi2020training} are not suitable for Anytime Inference, because each ensemble member has the exact same inference time. This results in a very limited set of possible inference times (see Figure~\ref{fig:motivation}). 
We also significantly outperform the tree-based ensemble HNE \cite{ruiz2021anytime} on both datasets, as well as the early-exit model MSDNet \cite{huang2017multi}. 
Lastly, we use \OurMethodSpace to construct a dynamic width and depth network. Our main comparison is to adaptation of Shapeshifter Networks \cite{plummer2020shapeshifter} to ensemble dynamic widths and depths.
We show a consistent improvement over Shapeshifter Networks across inference times. These results demonstrate that \OurMethodSpace can share parameters across members of diverse architectures more effectively than other approaches.
\smallskip

\noindent\textbf{Comparison of SuperWeight Clustering methods.}
Table \ref{table:cluster_assignment} demonstrates the effectiveness of our SuperWeight Clustering approach described in Section~\ref{subsec:learn_super}.  We provide four baselines: \emph{Shared Coefficients}, which learns SuperWeight Clusters, but shares coefficients between all layers (\ie, removing Section~\ref{subsec:learn_super}); \emph{Single SuperWeight Cluster}, which allows layers to have their own coefficients, but does not learn clusters (\ie, removing Section~\ref{sec:search}); \emph{Depth-binning}, a heuristic where we group together layers of the same relative depth across network architectures; and \emph{Coefficient Clustering} from prior work \cite{plummer2020shapeshifter}, which clusters coefficients $\alpha$ in Eq.~(\ref{eq:mixture}) to group layers.  We show that our approach outperforms these baselines.  Notably, we find that Coefficient Clustering performs in par or worse than other baselines.  In contrast, our gradient analysis approach (Section~\ref{subsec:learn_super}) takes into account the direction of change rather than just the current coefficient value. Thus, we obtain a 2\% gain on individual models and a small boost to ensembling performance with our approach (Table \ref{table:cluster_assignment}).  We show a visualization of SuperWeight cluster assignment in the supplementary.

\section{Conclusion}
We introduce \OurMethod, a method for learning parameter sharing patterns in single models as well as model ensembles. 
Our automatic sharing improves single model performance by up to 4\% compared to the baselines (Section \ref{sec:single_model}). \OurMethodSpace also match performance of efficient ensembles in the low-parameter regime, compared to prior work (Section \ref{sec:efficient}). When we add parameters, we outperform even deep ensembles on CIFAR with 17\% fewer parameters (Section \ref{sec:efficient}). Finally, \OurMethodSpace enables effective anytime inference (Section \ref{subsec:anytime_results}). We believe that \OurMethodSpace are a promising step forward in parameter-efficiency. Future work will include more deeply exploring architecture diversity; \cite{gontijo2021no} show that model architecture heterogeneity can be  key  to ensemble diversity on challenging tasks. 

\noindent\noindent\textbf{Broader Impacts and Limitations.} Effective parameter sharing allows one to use less compute, potentially running networks in efficient modes and conserving energy. However, it can also be used to maximize the use of compute if it's available, using \textit{more} energy with corresponding drawback. We urge readers to be aware of the carbon and energy footprint of the models they train. 

Although learning the sharing pattern (SuperWeight Clusters and coefficient sharing) is relatively lightweight, it does add computation to the learning process. Nevertheless, the improved predictive performance makes this a reasonable trade-off.

 \noindent\textbf{Acknowledgements}
This material is based upon work supported, in part, by DARPA under agreement number HR00112020054. Any opinions, findings, and conclusions or recommendations are those of the author(s) and do not necessarily reflect the views of the supporting agencies.

{\small
\bibliographystyle{ieee_fullname}
\bibliography{refs}
}

\newpage
\appendix

\section{Implementation Details}
\label{app:impl}

We provide the training details of both CIFAR~\cite{cifar100} and ImageNet~\cite{deng2009imagenet} experiments below.

\noindent\textbf{CIFAR.}  For CIFAR-10 and CIFAR-100 experiments, we train for 200 epochs with an initial learning rate of $0.1$. We use SGD with a Nestorov momentum value of 0.9. We use a weight decay value of $5e{-4}$ on all parameters  except weight coefficients. We decay the learning rate by a factor of $0.2$ at 60, 120, and 180 epochs. We use a batch size of 128, and use asynchronous BatchNorm across two devices (so BatchNorm batch size is 64). We pad images by 4 pixels and crop to 32 x 32 pixels, and also randomly flip and normalize such that it is zero-mean for training. We use the WideResNet 28-10 \cite{zagoruyko2016wide} architecture for all CIFAR experiments.
\smallskip

\noindent\textbf{ImageNet.} For ImageNet NPAS experiments we use a WideResNet 50-2 \cite{zagoruyko2016wide} architecture. We train for 90 epochs with a learning rate of 1.6. We use SGD with non-Nestorov momentum value of 0.9. We use a batch size of 1024. We decay the learning rate at epochs 30, 60, and 80 by a factor of 0.1. We use a label smoothing value of 0.1, and a weight decay of 0.0001. The weight decay is not applied to batch norm parameters. We do a linear warmup for the first 10 epochs. We use standard Inception-style data augmentation. For ensembleing experiments we use 64 16GB NVIDIA V100 GPUs, whereas for NPAS experiments we use 4 48GB NVIDIA A40 GPUs.

\subsection{Efficient Ensemble Experiments}

\noindent\textbf{Homogeneous Ensembles:}
We use WRN-28-10 models for all of our homogeneous ensembling experiments. For these models, we found that each layer having it's own SuperWeight works well, so did not perform the refinement step.
\smallskip

\noindent\textbf{Heterogeneous Ensembles:}
We train all models using the settings above except we train on a single GPU, the learning rate is decayed at epochs 60, 120, and 160, and we apply cutout \cite{devries2017improved} during training (unlike efficient ensembling using homogeneous ensembles, which does not use cutout for a fair comparison to prior work). \OurMethodShort-Multi-Width is a 3 member WideResNet (WRN) 28-[7,4,3] ensemble.  \OurMethodShort-Multi-Depth/Width is a 4 member WRN 28-[7,4] 16-[7,4] ensemble. All models start from 4 initial depth-binned SuperWeight Clusters and are refined using the gradient similarity threshold from Eq.\ (3), $\tau=0.1$. When learning where to share coefficients, SWE-Multi-Width is trained with the gradient similarity threshold from Eq.\ (2), $\beta=0.9$ for CIFAR-100 and $\beta=0.95$ for CIFAR-10. \OurMethodShort-Multi-Depth/Width uses $\beta=0.9$ for CIFAR-100 and $\beta=0.5$ for CIFAR-10.  See Section~\ref{sec:sensitivity} for a sensitivity study for hyperparameters $\beta$ and $\tau$.
\smallskip



\subsubsection{Baselines}
ShapeShifter Networks \cite{plummer2020shapeshifter}, Slimmable Networks \cite{yu2018slimmable}, Universally Slimmable Networks \cite{yu2019universally} are each trained using the same training settings as the SWE-Multi-Width and SWE-Multi-Depth/Width models (including using cutout for data augmentation). 

\noindent\textbf{Slimmable Networks} train a Multi-Width network by executing a subset of the channels for a predefined set of width configurations, called switches.
\smallskip

\noindent\textbf{Universally Slimmable Networks} extend Slimmable Networks to execute any width of a network within a max and minimum by training at each iteration a random subset of switches and calibrating the the batch normalization statistics of the final switches following training. We train Universally Slimmable using a max width of 7, a minimum width of 3, and the number of widths trained at each iteration, $n=4$. We calculate batch normalization statistics of network widths [3,4,5,6,7] over one epoch following training.
\smallskip

\noindent\textbf{ShapeShifter Networks (SSNs)} automatically learn where to share parameters within a network through clustering coefficients $\alpha$ to form groups and then within each group generating weights for layers from parameter banks via template mixing methods. For SSNs we give each ensemble member its own independent coefficients $\alpha$ and batch normalization parameters. For the Multi-Depth/Width network the layers are grouped manually by depth. 
\smallskip

\noindent\textbf{Multi-Scale Dense Network (MSDN)} \cite{huang2017multi} is a Multi-Depth network that provides dense connectivity between early exits and operates at multiple scales for an efficient Anytime Inference model. We train MSDNet using the training scheme from the original paper. All model hyperparameters are kept the same as the original except the network is built with 16 blocks with 12, 24, and 48 output channels for each of the three scales, respectively, in order to match the inference time with other methods. 
\smallskip

\noindent\textbf{Hierarchical Neural Ensemble} \cite{ruiz2021anytime} is a Multi-Depth network that trains a tree based ensemble using a novel distillation loss. Results are pulled directly from the original paper with models evaluated for inference time using the author's code. The model uses a modified ResNet-50 with $N=16$ ensemble members.
\smallskip

\noindent\textbf{Efficient Homogeneous Ensemble Baselines}: BatchEnsemble \cite{wen2020batchensemble} and MIMO \cite{havasi2020training} are two homogeneous ensembling methods (which we apply to the anytime inference task in Figure 1 of the main paper). BatchEnsemble perturbs weights with rank-1 matrix for each ensemble member. MIMO hard shares all parameters between ensemble members, except for the input convolutional layer and output layer. BatchEnsemble is ineffective as an anytime inference method because there are only $N$ evaluation speeds, where $N$ is the number of ensemble members. MIMO only provides a single inference speed because of its shared backbone. 

\renewcommand{\algorithmicrequire}{\textbf{Input:}}

\begin{algorithm}[ht]
\caption{Priority-Queue Group Assignment}\label{alg:priority_assignment}
\begin{algorithmic}
\Require threshold $\epsilon$, priority-queue $Q = \{\text{gradient similarity } \psi_{i,j}: (\text{layers } \ell_i,\ell_j) \}$ 
\\
\State Groups $G = list()$
\While{$|Q| > 0$}
    \State $\psi_{i,j}, \ell_i,\ell_j = Q.pop()$

    \If{ $\psi_{i,j} > \epsilon$ }
        \If{ $\ell_i \in g \And \ell_j \in g'$, where $g,g' \in G$ }
        \State $G.append(g \cup g')$  \Comment{Merge into new group}
        \State delete $g, g'$ from $G$ 
        \ElsIf{ $\ell_i \in g$ } \Comment{One layer already belongs to a group}
        \State $g = g \cup \{\ell_j\}$
        \ElsIf{ $\ell_j \in g'$} 
        \State $g' = g' \cup \{\ell_i\}$
        \Else \Comment{Neither layer belongs to a group}
        \State G.append(\{$\ell_i, \ell_j$\})
        \EndIf
    \Else \Comment{similarity is below threshold, so add any remaining ungrouped layers}
        \If{ $\ell_i \notin g, \forall g \in G$ } 
        \State G.append(\{$\ell_i$\})
        \EndIf
        \If{ $\ell_j \notin g', \forall g' \in G$ } 
        \State G.append(\{$\ell_j$\})
        \EndIf
\EndIf
\EndWhile
\\
\\
\Return $G$

\end{algorithmic}

\end{algorithm}

\section{Priority-Queue Weight Template Assignment}
\label{app:pq}

In Section 2.2 of our main paper we introduce a method that uses gradient similarity between shared parameters in order to determine where sharing is effective.  The intuition behind our approach is that sharing between layers is likely less effective when those layers provide conflicting gradients to the shared parameters.  We proposed a greedy approach that would train a model for $N$ epochs and then measure similarity between gradients supplied to the shared parameters aggregated over an entire epoch.  Gradient similarity between layers $\ell_i, \ell_j$, which we denote as $\psi_{i,j}$, is computed using Eq.\ (2) when learning where Weight Templates should share coefficients, and Eq.\ (3) when learning SuperWeight Clusters.  This similarity would be used to construct a priority-queue $Q$, which we traverse merging any layers into a group $G$ that are above some threshold $\epsilon$, with any remaining layers being placed into the same group (see Algorithm \ref{alg:priority_assignment}).  Note that in our experiments $\epsilon$ is a threshold on gradient similarity, but one could also use a threshold on the number groups instead (or in combination), which we leave for future work. Layers in the same group would continue to share parameters, whereas layers in different groups would no longer completely share parameters.

\section{Additional Efficient Ensembling Experimental Results}

\subsection{Additional Efficient Ensemble Results for Anytime Inference}

\begin{figure}[t]
\centering
\includegraphics[width=0.48\textwidth]{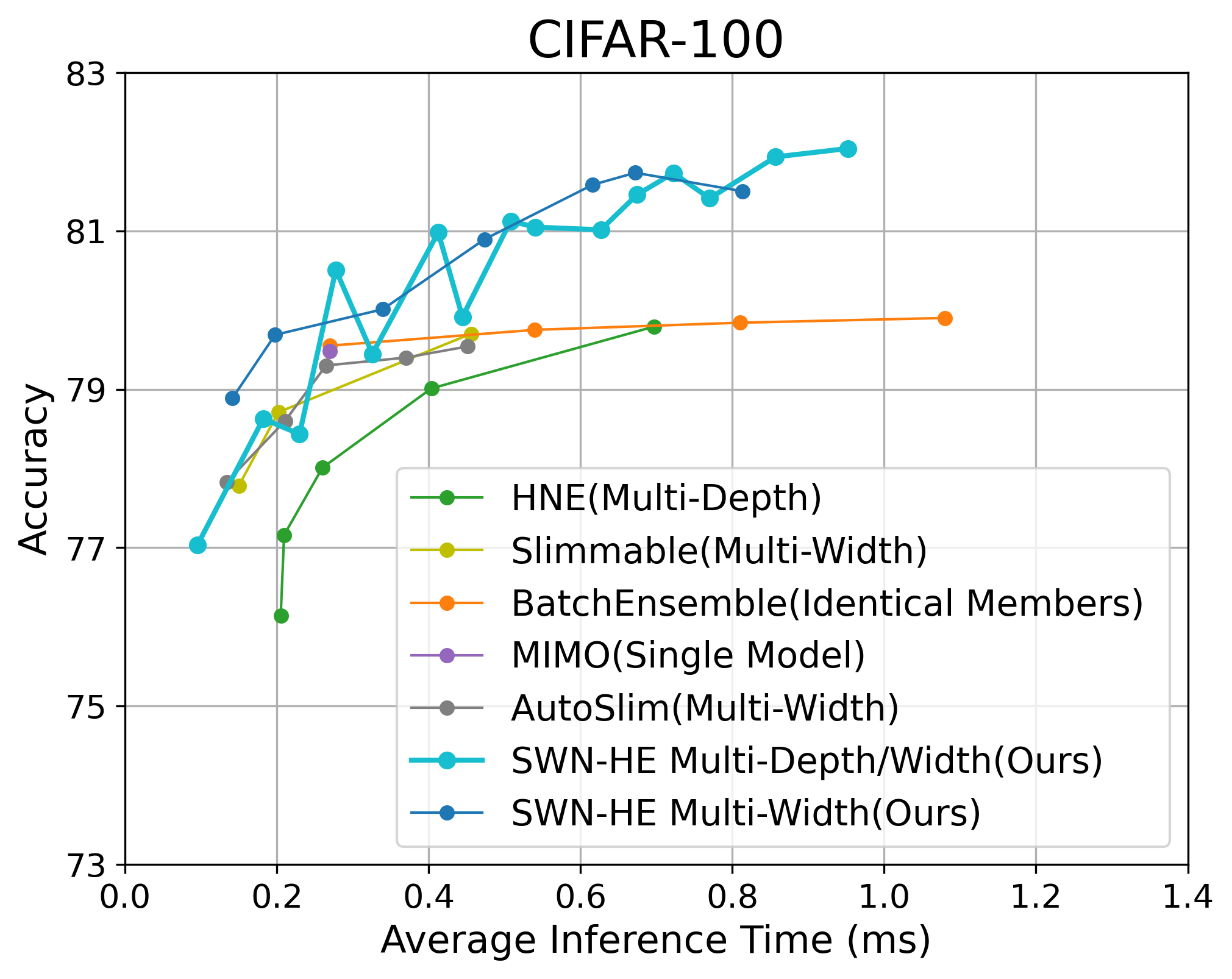}
\caption{\OurMethod learns effective soft parameter sharing between members, even for diverse architectures.   This enables our approach to support a range of inference times while outperforming prior work in efficient ensembling and anytime inference~\cite{ruiz2021anytime,havasi2020training,wen2020batchensemble,yu2018slimmable,yu2019autoslim} on CIFAR-100 using WRN-28-5~\cite{zagoruyko2016wide}. Additionally, some efficient ensembling methods like MIMO~\cite{havasi2020training} do not enable multiple inference times at all, represented  as a point on the figure. }
\label{fig:efficient_ensemble_anytime}
\end{figure}

 Prior work in efficient ensembling (\eg,~\cite{lee2015m,wen2020batchensemble,wenzel2020hyperparameter}) uses hand-crafted strategies that required ensemble members to have identical architectures and adds diversity by perturbing weights and/or features.  In contrast,  our \OurMethod, which learn effective soft parameter sharing between members, even for diverse architectures.  As shown in Figure \ref{fig:efficient_ensemble_anytime}, this enables our approach to support a range of inference times while outperforming prior work in efficient ensembling and anytime inference~\cite{ruiz2021anytime,havasi2020training,wen2020batchensemble,yu2018slimmable,yu2019autoslim} on CIFAR-100 using WRN-28-5~\cite{zagoruyko2016wide}. Additionally, some efficient ensembling methods like MIMO~\cite{havasi2020training} do not enable multiple inference times at all.


\subsection{Slimmable Networks and Universally Slimmable Networks Ensemble Comparison}

\begin{figure*}[t]
    \centering
    \includegraphics[width=1.0\textwidth]{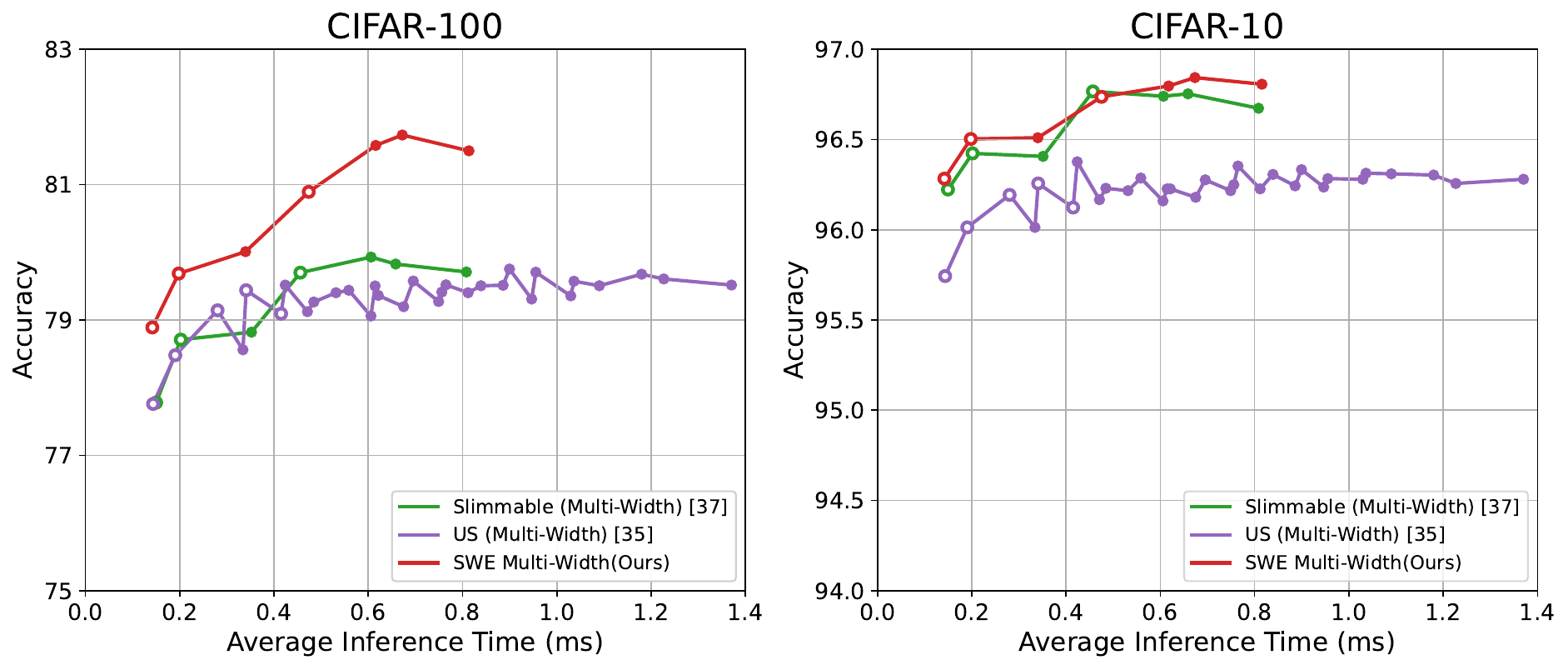}
    \caption{\textbf{Slimmable Networks and Universally Slimmable Networks ensemble comparison.} Anytime inference results using Slimmable \cite{yu2018slimmable} and Universally Slimmable \cite{yu2019universally} as ensembles of multiple widths compared to our \OurMethodShort-Multi-Width.}
    \label{fig:slimmable_ensemble}
\end{figure*}

Slimmable Networks \cite{yu2018slimmable} and Universally Slimmable Networks (US) \cite{yu2019universally} both train a dynamic width network such that given a budget at inference time, one can match the budget by running a slim version of the network through executing a subset of the channels at each layer. These methods can be run as an ensemble similar to ours by running inference through multiple widths. In Figure \ref{fig:slimmable_ensemble} we compare our \OurMethodShort-Multi-Width model to Slimmable and Universally Slimmable ensembles. SuperWeight Ensembles benefit from ensembling diverse architectures, improving performance.  In contrast, ensembling multiple widths of Slimmable and Universally Slimmable Networks only leads to a slight boost, or even decrease, in accuracy.

\subsection{CIFAR-10-C}

We present results on CIFAR-10-C below. Compared to efficient ensembling baselines, our \OurMethodShort-HE outperforms others on two out of three metrics, while also providing flexibility not present in prior work via our heterogeneous ensembles and the ability to adjust the number of parameters in our network without changing architecture.

\begin{table}[t]
\centering
 \begin{tabular}{l ccc} 
 \hline
 Method & Top-1 Acc  &  NLL & ECE  \\ 
 \hline
 { BatchEnsembles  \cite{havasi2020training}} & \textbf{77.5}  &  1.02 & 12.9\\
 { MIMO \cite{havasi2020training}} & 76.6  &  0.927 & 11.2  \\
 { \OurMethod (\textbf{ours})} & 76.01 &  0.885 & 10.2\\ 
 { \OurMethod (\textbf{ours})}  & 76.6 &  \textbf{0.872} & \textbf{8.8} \\
 \hline
\end{tabular}
\caption{Efficient ensembling comparison on CIFAR-10-C}
\label{table:cifar10c}
\end{table}

\subsection{Ensemble Diversity Analysis}
\label{diversity}


A key attribute of ensembles which makes them effective is their diversity; if the errors of ensemble members are not decorrelated then there is no additional benefit of doing inference through additional ensemble members. In this section, we demonstrate that \OurMethodSpace accomplish this. 

We use a diversity metric introduced in Fort et al.~\cite{fort2019deep}, which measure the fraction of differing predictions by two ensemble members, normalized the by the error of one of them. We present these results in Table \ref{table:diversity}; we can see that \OurMethodSpace is more diverse than all other shared parameter methods. One interesting finding is that  120M parameter \OurMethodSpace  outperform Standard Deep Ensembles, yet have lower diversity. Looking deeper, the individual model accuracies are improved for \OurMethodSpace (average deep 79.9\% vs average \OurMethodSpace 80.4\%) Therefore, it seems like \OurMethodSpace helps the model generalize better. This could come from the fact that the parameter factorization limits the space of possible weights. This is especially interesting because WRN-28-10 is already a highly optimized model from a hyper-parameter perspective. Of note is that one of the ensemble members from the \OurMethodSpace gives higher performance (80.5\% for the best model) than a standard model (80.1\%). Therefore one could use \OurMethodSpace to train a highly performant single model, which signifies another advantage our approach has over prior work~\cite{havasi2020training,wen2020batchensemble}.

\begin{figure*}[t]
\centering
\includegraphics[width=\textwidth]{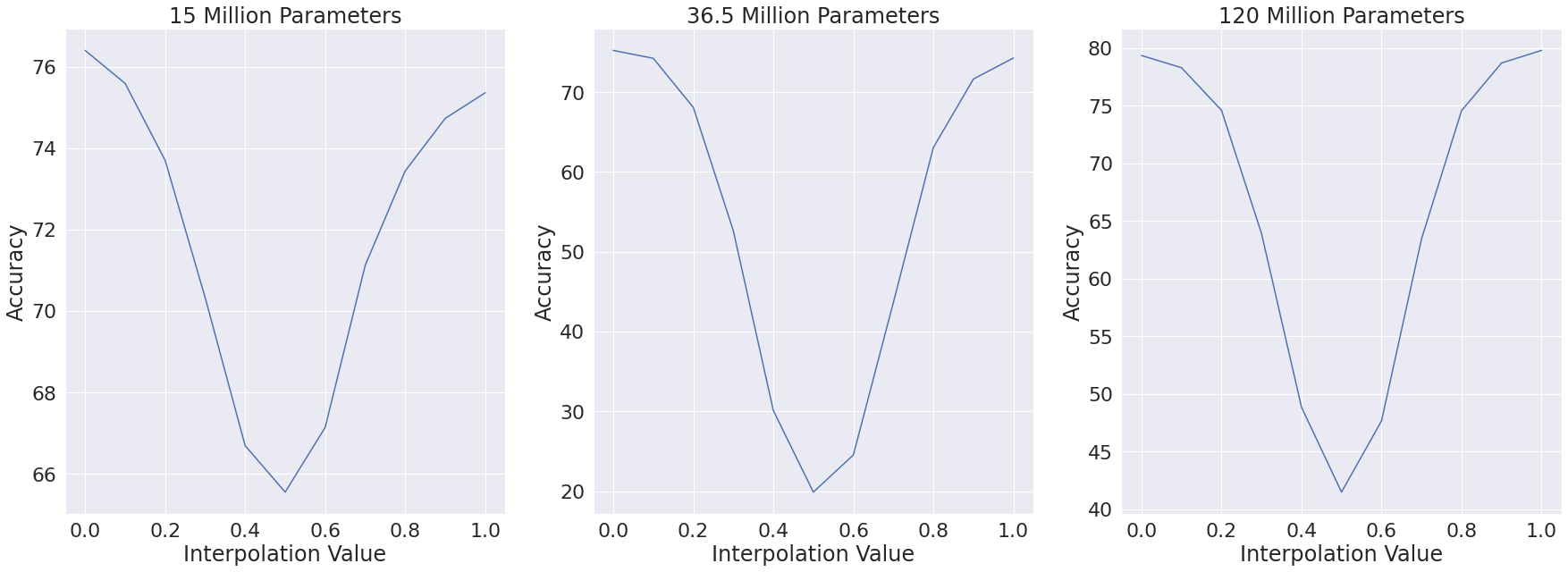}
\caption{Linear interpolations in parameter space on CIFAR-100, with different numbers of parameters in an ensemble of size 4. We plot accuracy vs interpolation point. Because the accuracy dips, we know the ensemble members are diverse and find different local minima. }
\label{fig:interpolations}
\end{figure*}

To provide another diversity metric, in Figure \ref{fig:interpolations} we interpolate between two WRN-28-10 ensemble members in parameter space to see if the models indeed are in different optimization basins, and report accuracy at each operating point. We accomplish this by interpolating parameters, and leaving Batch Normalization \cite{ioffe2015batch} in train mode because accumulated statistics are not meaningful at interpolated points.  If the interpolates have high accuracy, this  indicates the ensemble members landed in the same optimization basin and therefore are not as diverse as they could be. We find that interpolates have much decreased accuracy compared to the end points, supporting the idea that our model learns diverse ensemble members. 

We can see that although all networks experience some degree of accuracy drop between ensemble members,  the accuracy drop for the low-parameter model is significantly lower. This seems to indicate that as the function space becomes constrained with a lower number of parameters in the parameter bank, the ensemble diversity starts to suffer.

\begin{table}[t]
\centering
 \begin{tabular}{l cc} 
 \hline
 Method & Params  &  Diversity  \\ 
 \hline
 { Standard Deep Ensemble
  \cite{havasi2020training}} & 146.0M &  0.88 \\
 { BatchEnsembles  \cite{havasi2020training}} & 36.5M &  0.40\\
 { MIMO \cite{havasi2020training}} & 36.5M &  0.91 \\
 { \OurMethodSpace (\textbf{ours})} & 36.5M &    0.78\\ 
 { \OurMethodSpace (\textbf{ours})}  & 120.0M &  0.85 \\
 \hline
\end{tabular}
\caption{\textbf{Diversity on CIFAR-100.} Diversity is measured as the proportion of samples two ensemble members disagree on, normalized by the error rate of one of the member as in \cite{fort2019deep}.  \OurMethodSpace are more diverse than BatchEnsembles, the other shared parameter model. Although our 120M model is slightly less diverse than Standard Deep Ensembles, the performance is higher, due to the average member accuracy being higher than Standard Deep Ensembles}
\label{table:diversity}
\end{table}
\begin{figure*}[t]
    \centering
    \includegraphics[width=1.0\textwidth]{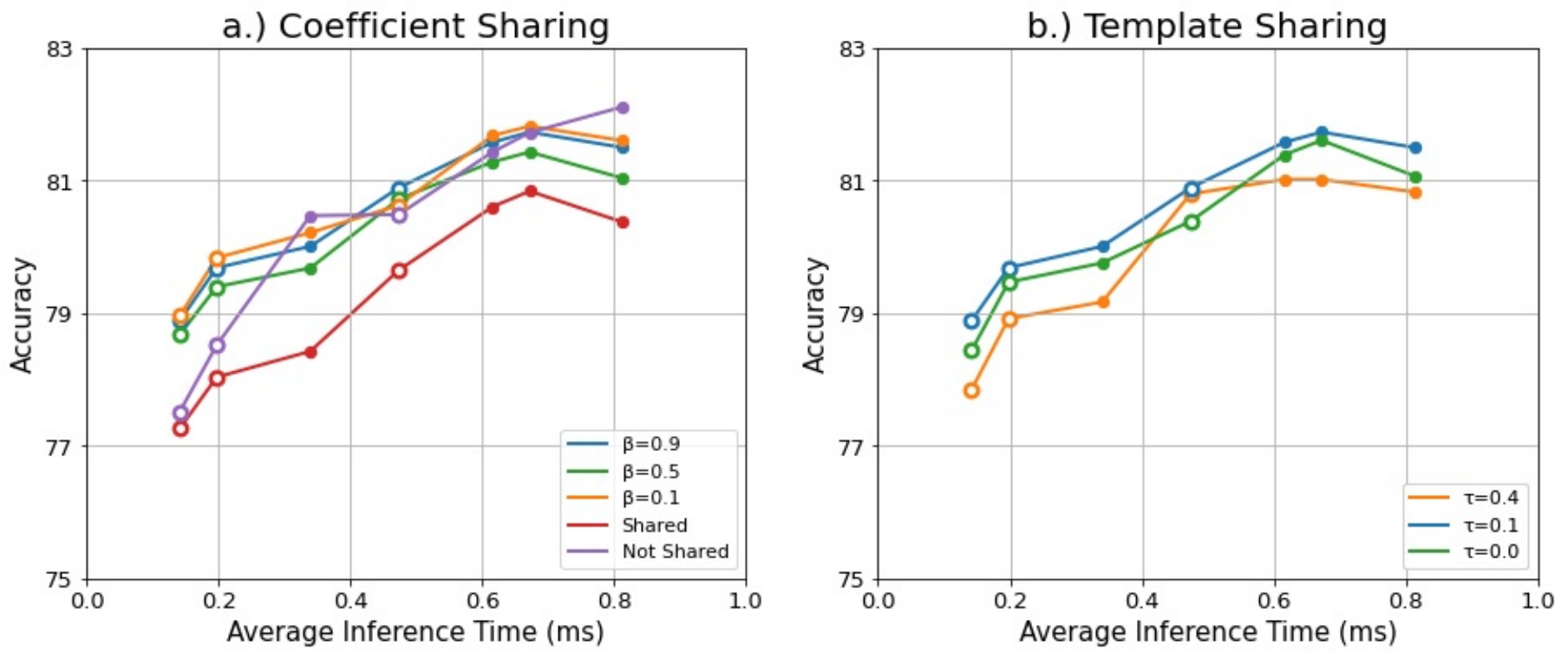}
    \caption{\textbf{Gradient analysis hyperparameter sensitivity.} Anytime inference results on CIFAR-100 \cite{cifar100} using \OurMethodShort-Multi-Width architecture when: \textbf{a)} varying the gradient similarity threshold $\beta$ from Eq.\ (3) for coefficient sharing, no coefficient sharing, and strict coefficient sharing; and \textbf{b)} varying the gradient similarity threshold $\tau$ from Eq.\ (2) for SuperWeight Cluster forming. For both $\beta$ and $\tau$ best performance is between 0.0 and 1.0, indicating that too much sharing and too little sharing are both harmful.}
    \label{fig:hyperparam_sensitivity}
\end{figure*}

\subsection{Gradient Analysis Hyperparameter Sensitivity}
\label{sec:sensitivity}

SuperWeight Ensembles learn where to share parameters through two steps of gradient analysis; first to form SuperWeight Clusters (Section 2.2.1), and then to separate SuperWeights by decoupling shared coefficients (Section 2.2.2). Here we explore the sensitivity of the hyperparameters used in each step.

When separating SuperWeights using Eq.\ (3), we give unique coefficients to SuperWeights where the similarity between the gradients of shared coefficients is less than $\beta$. $\beta$ is used as the threshold $\epsilon$ in Algorithm \ref{alg:priority_assignment} for separating coefficients. 
In Figure \ref{fig:hyperparam_sensitivity}(a) we show how sensitive our method is to the selection of $\beta$ using the \OurMethodShort-Multi-Width architecture on CIFAR-100 \cite{cifar100}. In addition to reporting results for values of $\beta \in \{0.1,0.5,0.9\}$ (note that $\tau=0.1$), we also report results when no coefficients are shared, and when all coefficients are shared between layers sharing Weight Templates. The results in Figure \ref{fig:hyperparam_sensitivity}(a) show that no-coefficient and strict coefficient sharing (referred to as ``Not Shared'' and ``Shared,'' respectively) underperform compared to using our approach.  Note that we found optimal values of $\beta$ to come from dataset-specific tuning. 

When generating SuperWeight Clusters using Eq.\ (2) we split layers into separate groups if the gradient similarity is less than $\tau$. The higher $\tau$, the more groups are formed, the less layers sharing Weight Templates, and the less parameters given to each Weight Template. Setting $\tau \geq 0$ results in sharing between all layers which have gradients that are not conflicting on average.  As $\tau$ is increased, layer gradients must point in closer directions to be shared.  Note $\tau$ is given as the input value to $\epsilon$ in Algorithm \ref{alg:priority_assignment}. In Figure \ref{fig:hyperparam_sensitivity}(b) we report best performance comes when $\tau=0.1$ when using the \OurMethodShort-Multi-Width architecture on CIFAR-100 \cite{cifar100}, which we found to be consistent across settings and datasets.  Thus, we use this same value of $\tau$ across all experiments.

\subsection{Performance under severe parameter constraint}

One key feature of \OurMethodSpace is that the parameter count is decoupled from backbone. It is therefore interesting to see how the network behaves under stronger parameter constraints. In Table \ref{table:parameters}, we present CIFAR-100 top-1 accuracies under various parameter constraints. \OurMethodShort-Single Superweight refers to the homogeneous ensemble which has a single superweight per layer. This is what is presented in Table 2 of our paper. \OurMethodShort-Gradient Conflict is also a homogeneous ensemble, but with learned SuperWeight Clusters using the gradient conflict criterion. Finally, \OurMethodShort-Heterogeneous is a  WRN 34-8, 28-12, 28-10, and 28-8 ensemble. The gradient conflict criterion becomes more important at lower parameter counts. The improvement of the heterogeneous ensemble over the homogeneous ensemble also increases with decreased parameter counts.

\begin{table}[t]
\centering
 \setlength{\tabcolsep}{2pt}
 \begin{tabular}{lccc} 
 \hline
 Method & 7 million  &  15 million & 36.5 million  \\ 
 \hline
{\OurMethodShort-Single Superweight}  & 79.2\%    & 81.3\%    & 82.3\%       \\ 
{\OurMethodShort- Gradient Conflict} & 80.1\%   & 81.4\%    & 82.2\%      \\ 
{\OurMethodShort- Heterogenous}       & 80.8\%   & 81.6\%    & 82.4\%       \\
\hline
\end{tabular}
\caption{\textbf{Changing parameter constraint:} CIFAR-100 Top-1 accuracies. The gradient conflict criterion becomes more important at lower parameter counts. The improvement of the heterogenous ensemble over the homogenous ensemble also increases with decreased parameter counts. }
\label{table:parameters}
\end{table}

\subsection{SuperWeight Cluster Membership}
\label{app:cluster_membership}

\begin{figure}[t]
    \centering
    \includegraphics[width=1.0\columnwidth]{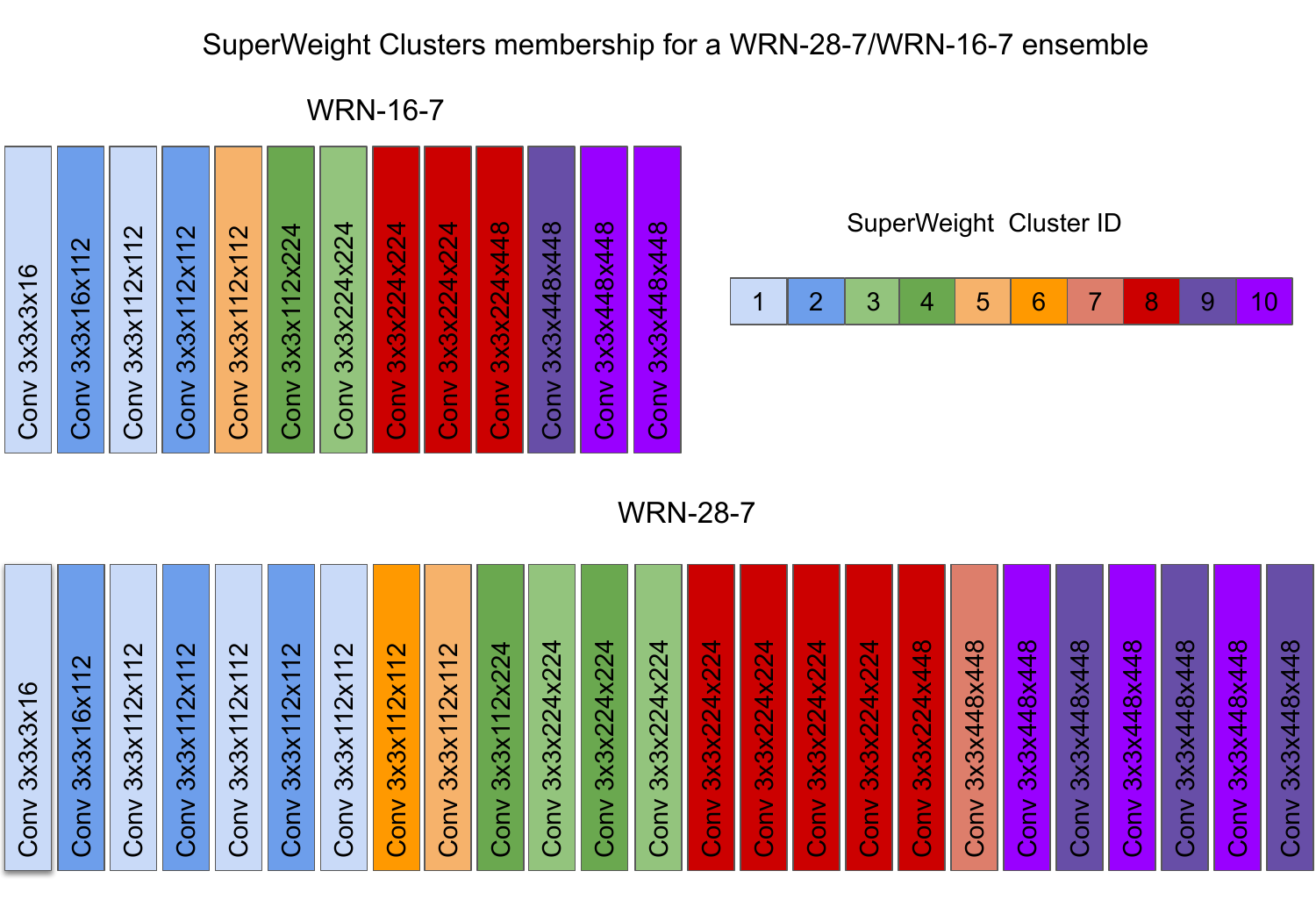}
    \vspace{-4mm}
    \caption{\textbf{SuperWeight Cluster Membership} of a two member ensemble, a WRN-28-7 and WRN-16-7. The initial depthwise sharing pattern would segment each network into four equal groups of layers. Our gradient-based learning of SuperWeight clusters allows indvidual layers to specialize. Interestingly, it seems like our method learns to separate adjacent layers into separate SuperWeight clusters. This could increase diversity between ensemble members.  }
    \label{fig:cluster_membership}
\end{figure}

In Figure \ref{fig:cluster_membership}, we present sharing patterns (SuperWeight Cluster membership) of two networks (WRN-28-7/WRN-16-7). The initial depthwise sharing pattern would segment each network into four equal groups of layers. Our gradient-based learning of SuperWeight clusters allows indvidual layers to specialize. Interestingly, it seems like our method learns to separate adjacent layers into separate SuperWeight clusters. This could increase diversity between ensebmle members.  

\subsection{Diverse Architecture Families}

Although we show results primarily on ResNets in our work, our sharing methodology does indeed function well for diverse ensembles. For example, consider an ensemble consisting of a Mobilenetv2 and WRN-28-5, shown in Table \ref{tab:multi_arch}. We can see that our sharing procedure even helps compared to standard Deep Ensemble performance, despite diverse architectures. Note that these results are over 2 ensemble members, whereas the results in Table 2 of our paper use 4 ensemble members.

\begin{table}[t]
\centering
 \begin{tabular}{lccc} 
 \toprule
 Method   & Top-1 & NLL & ECE  \\ 
 \midrule
Deep Ensembles &   80.1\% & 0.776 & 6.5\%\\
\midrule
\OurMethodShort-HE &   80.3\% & 0.762 & 5.3\%\\
\bottomrule
\end{tabular}
\caption{\textbf{Diverse Architecture Families:} We compare the performance of an ensemble consisting of a Mobilenetv2 and WRN-28-5 to standard deep ensembles. We see that our sharing procedure boosts performance despite diverse architectures. 
 }
\label{tab:multi_arch}
\end{table}

\begin{table*}[t]
\centering
\setlength{\tabcolsep}{1.3pt}
{\footnotesize
 \begin{tabular}{rlcccccccccc} 
 \hline
  & & & \multicolumn{3}{c}{CIFAR-100 (clean)} & \multicolumn{3}{c}{CIFAR-100-C} & \multicolumn{3}{c}{CIFAR-10}\\ 
  \hline
 & Method & Params & Top-1 $\uparrow$ &  NLL $\downarrow$ &  ECE $\downarrow$ & Top-1 $\uparrow$ & NLL $\downarrow$ & ECE $\downarrow$ & Top-1 $\uparrow$ & NLL $\downarrow$ & ECE $\downarrow$\\
 \hline
  \textbf{(a)} & {WRN-28-10 }& 36.5M & 79.8 & 0.875 & 8.6 & 51.4 & 2.70 & 23.9 & 96.0 & 0.159 &  2.3\\
 & {BE  }& 36.5M & 81.5 & 0.740  & 5.6 & 54.1 & 2.49 & 19.1 & 96.2 & 0.143 & 2.1\\
 & {BE + EnsBN} & 36.5M & 81.9 & n/a & 2.8 & 54.1 & n/a & 19.1 & 96.2 & n/a & 1.8\\
 & {MIMO  }& 36.5M & 82.0 & 0.690  & 2.2 & 53.7 & 2.28 & 12.9 & \textbf{96.4} & 0.123 & 1.0\\
 & {\OurMethodShort-HO (\textbf{Ours})} & 36.5M & 82.2 $\pm$ 0.28 & 0.702 $\pm$ 0.009 & 2.7 $\pm$ 0.04  & 52.9 $\pm$ 0.24 & \textbf{2.17 $\pm$ 0.02} & 10.3 $\pm$ 0.47 & 96.3 $\pm$ 0.05 & 0.120 $\pm$ 0.002 & \textbf{0.8 $\pm$ 0.08}\\
 & {\OurMethodShort-HE (\textbf{Ours})} & 36.5M & \textbf{82.4 $\pm$ 0.02} & \textbf{0.663 $\pm 0.001$} & 3.0 $\pm$ 0.23 & 53.0 $\pm$ 0.06 & \textbf{2.17 $\pm$ 0.01} & \textbf{10.0 $\pm$ 0.45} & \textbf{96.5 $\pm$ 0.02} & \textbf{0.115 $\pm$ 0.003} & \textbf{0.8 $\pm$ 0.03}\\
 \hline
  \textbf{(b)} & {Deep Ensembles } & 146M & 82.7 & \textbf{0.666} & \textbf{2.1} & 54.1 & 2.27 & 13.8 & \textbf{96.6} & \textbf{0.114} &  1.0\\
  & {\OurMethodShort-HO (\textbf{Ours}) }& 120M &{ } \textbf{82.9 $\pm$ 0.05} &{ } \textbf{0.666 $\pm$ 0.007} &{ } 2.2 $\pm$ 0.3 &{ } \textbf{54.7 $\pm$ 0.05} &{ } \textbf{2.00 $\pm$ 0.01} &{ } \textbf{10.3 $\pm$ 0.01} &{ } \textbf{96.6 $\pm$ 0.09} &{ } 0.119 $\pm$ 0.001 &{ }  \textbf{0.8 $\pm$ 0.07}\\
 \hline
\end{tabular}
}
\caption{\textbf{Homogeneous ensembling comparison} on CIFAR-100 (clean)~\cite{cifar100} CIFAR-100-C (corrupt)~\cite{hendrycks2019robustness}, and CIFAR-10~\cite{cifar100} using the WideResNet architecture~\cite{zagoruyko2016wide} averaged over three runs. We add error bars representing standard error for our experiments for context. Prior work did not report error bars. \textbf{(a)} shows that our approach outperforms prior work in efficient ensembling. \textbf{(b)} compares the performance of increasing the number of parameters (without changing the architecture) using our approach compared to standard Deep Ensembles, which trains 4 independent networks as ensemble members.
}
\label{table:homogenous}
\end{table*}

\subsection{Efficient Ensembles on CIFAR}
\label{appendix:error_bar_section}
In Table \ref{table:homogenous} we present CIFAR results from the main paper, with error bars (representing standard error) to provide additional context. Note that even with this additional information, it is clear our method outperforms all baselines on CIFAR in the low parameter regime, and even outperforms standard ensembles in the high parameter regime (with 17\% fewer parameters). Results we report for baselines are taken from prior work and do not provide error bars.

\end{document}